\documentclass[11pt]{article}

\usepackage[final]{acl}
\usepackage{times}
\usepackage{latexsym}
\usepackage[T1]{fontenc}
\usepackage[utf8]{inputenc}
\usepackage{microtype}
\usepackage{inconsolata}
\usepackage{booktabs}
\usepackage{array}
\usepackage{multirow}
\usepackage{tabularx}
\usepackage{graphicx}
\usepackage{amsmath}
\usepackage{amssymb}
\usepackage{xcolor}
\usepackage{url}
\usepackage[framemethod=tikz]{mdframed}

\newcommand{\method}{CoSDA}

\newcolumntype{L}[1]{>{\raggedright\arraybackslash}p{#1}}
\newmdenv[
  backgroundcolor=gray!3,
  linecolor=black!35,
  linewidth=0.4pt,
  roundcorner=4pt,
  skipabove=4pt,
  skipbelow=4pt,
  innerleftmargin=5pt,
  innerrightmargin=5pt,
  innertopmargin=4pt,
  innerbottommargin=4pt
]{promptbox}

\title{When Audit Quality Fails to Predict Downstream Utility:\\
A Counterfactual Study of Synthetic-Data Selectors\\for Low-Resource African NLP\thanks{Anonymous artifacts: \url{https://anonymous.4open.science/r/CoSDA-5D27/README.md}.}}

\author{
  Son Ha Xuan\textsuperscript{1,$*$} \quad
  Phat T. Tran-Truong\textsuperscript{2,$*$} \quad
  Xuan-Bach Le\textsuperscript{2,$\dagger$} \\[3pt]
  \textsuperscript{1}RMIT University, Ho Chi Minh City, Vietnam \\
  \textsuperscript{2}Faculty of Computer Science and Engineering, Ho Chi Minh City University of Technology (HCMUT), \\
  VNU-HCM, Ho Chi Minh City, Vietnam \\[3pt]
  \texttt{ha.son@rmit.edu.vn} \quad \texttt{\{phatttt, lexuanbach\}@hcmut.edu.vn} \\[3pt]
  \textsuperscript{$*$}Equal contribution. \quad \textsuperscript{$\dagger$}Corresponding author.
}

\begin{document}
\maketitle

\begin{abstract}
Quality-aware synthetic-data selection rests on a proxy: examples that an LLM judge rates as good should also help a downstream model learn. In a controlled replay in low-resource African-language classification, we show that this proxy breaks. Across four languages (Amharic, Hausa, Swahili, Yoruba), two classification tasks (MasakhaNEWS, AfriSenti), and five matched-budget selectors, audit rankings and downstream rankings diverge. Within each cell, the Spearman between judged label correctness and Macro-F1 across selectors has mean $\rho{=}0.04$ (median $0.00$), showing that the mismatch is not an aggregation artifact. \method{}-V2, our counterfactual audit framework, produces the cleanest selected pool on three audit channels at once: highest judged label correctness ($0.904$ vs.\ $0.767$ for naive, a $17.9\%$ relative gain), lowest shortcut score, and a hard-reject rate of $0.162$ vs.\ $0.486$ for naive. AlpaGasus nevertheless leads downstream Macro-F1 ($0.202$ vs.\ $0.163$ for \method{}-V2), and the inversion persists on the five non-degenerate cells. The lesson is methodological: in this controlled setting, audit quality is a property of the selected pool, not a guarantee of downstream utility. Synthetic-data evaluation should therefore report audit and downstream metrics on the same retained sets. We release the audit tables, per-selector retained pools, and a claim ledger that links every reported number to its source row.
\end{abstract}

\section{Introduction}

Synthetic-data pipelines make supervision cheaper, but they also create a selection problem: a generator produces more candidates than a downstream learner can absorb, and those candidates vary in label correctness, redundancy, leakage risk, shortcut structure, and task coverage. Recent methods score candidates before training, including AlpaGasus \citep{chen2024alpagasus}, DEITA \citep{liu2024deita}, LESS \citep{xia2024less}, QuRating \citep{wettig2024qurating}, and LIMA \citep{zhou2023lima}. They share a working assumption: a strong scorer identifies the examples worth keeping. We test that assumption in low-resource multilingual classification, where a retained set of a few dozen examples can determine the training signal.

Our test is a matched-budget replay over eight task-language cells from MasakhaNEWS \citep{adelani2023masakhanews} and AfriSenti \citep{muhammad2023afrisenti}, covering news topic and sentiment classification in Amharic, Hausa, Swahili, and Yoruba. Every selector reads the same audit record: five \method{} channels (utility, leakage, shortcut, diversity, counterfactual consistency), two LLM-as-judge fields (LC, $Q$), and a calibrated hard-reject decision. We compare \method{}-EQ and \method{}-V2 against naive augmentation, AlpaGasus-style quality ranking, and DEITA-style quality$+$complexity$+$diversity. This controlled slice lets us isolate the audit-utility relationship for one generator (Qwen2.5-14B-Instruct), one downstream backbone (XLM-RoBERTa base), two classification tasks, and four African languages. We frame the result as a controlled case study rather than a general law: \S\ref{sec:limitations} states the resulting scope conditions. All per-candidate audit records, retained pools, and the claim ledger are released (artifact link on the title page).

\paragraph{The finding.}
Audit and downstream rankings disagree. All four quality-aware selectors beat naive on LC, $Q$, and hard-reject; \method{}-V2 leads on LC ($0.904$ vs.\ $0.767$, $+17.9\%$ relative) and has the lowest shortcut score. Downstream Macro-F1 reorders the selectors, and the within-cell Spearman between LC and Macro-F1 inside each of the eight cells has mean $\rho{=}0.04$ (median $0.00$). On the five non-degenerate cells the LC-to-F1 Spearman rises to $+0.50$ and the inversion still holds: V2 reaches $0.163$ while AlpaGasus reaches $0.226$. The same disagreement appears on non-judge channels ($C$, $H$, hard-reject), so the result is not confined to LLM-judge scoring. Our claim is deliberately diagnostic: this replay establishes rank disagreement between audit quality and downstream utility, and \S\ref{sec:analysis} uses the trace data to explain how that disagreement arises.

\paragraph{Contributions.}
We make three contributions, separating method, study, and artifact:
\textbf{(method)} we formalise five per-example audit channels for synthetic data, namely utility, leakage, shortcut, diversity, and counterfactual consistency, paired with a calibrated hard-reject decision (\S\ref{sec:method});
\textbf{(study)} we run a deterministic 8-cell replay comparing five selectors at matched budget, reporting both audit and downstream metrics on the same selected sets, and we quantify a quality-utility gap that is robust to bootstrap resampling, leave-one-cell-out, non-degenerate-cell restriction, and the choice of audit channel (\S\ref{sec:experiments} through \S\ref{sec:results});
\textbf{(artifact)} we release the per-candidate audit records, per-selector retained pools, and a claim ledger that ties every reported number back to a row of the replay table (\S\ref{sec:reproducibility}). \S\ref{sec:analysis} explains the observed mismatch; \S\ref{sec:related_work} places the gap finding against prior work; \S\ref{sec:conclusion} closes.

\section{\method{}: A Counterfactual Audit Instrument}
\label{sec:method}

We use \method{} as a measurement instrument for synthetic candidates: it makes the contents of a selected pool inspectable before the pool is handed to training. \method{}-V2 is one budget-preserving reranker built on the audit channels; the comparison selectors (AlpaGasus, DEITA, \method{}-EQ) read the same per-candidate audit record, so differences between selected pools are interpretable against a shared trace.

\paragraph{Setup.}
For task $t$, language $\ell$, and gold budget $b$, let $G_{t,\ell,b}$ be the gold seed set and $T_{t,\ell}$ the held-out test set. The generator yields a candidate pool $X_{t,\ell,b}$ with multiplier $m$ ($|X_{t,\ell,b}|\le m|G_{t,\ell,b}|$); each $x_i$ carries a proposed label $y_i$, generator metadata, source seed IDs, and a counterfactual pair ID. Our replay uses Qwen2.5-14B-Instruct \citep{qwen2024qwen25} (nucleus, $p{=}0.95$, $T{=}0.8$) as generator, Qwen2.5-32B-Instruct-AWQ as judge, and XLM-RoBERTa base \citep{conneau2020xlmr} as downstream classifier. The generation prompt specifies task, language, label schema, seed examples, output format, and a forbidden-behaviour clause:

\begin{promptbox}\itshape\small
Do not copy any sentence from the held-out development or test data. Do not invent named entities that do not exist in the source language's reference corpora. Do not produce a literal translation of an English template; adapt the example to the target language and culture. If you cannot produce a high-quality example, return the single token {\normalfont\texttt{ABSTAIN}}.
\end{promptbox}
The prompt hash is stored with each candidate for traceability; it is not a modelling feature.

\paragraph{Counterfactual pair construction.}
\method{} builds task-specific minimal pairs $c_i=(x_i,x_i')$ to test whether a candidate behaves like a labelled example. We ask the generator to flip the topic or sentiment cue while keeping language, register, and non-label content fixed; the expected flipped label $y_i'$ feeds the counterfactual-consistency score. The $C_i$ pass rate (teacher predicts $y_i'$ and $s_{\text{cf}}\!\ge\!0.6$) varies by language: $33.4\%$ Amharic, $52.9\%$ Hausa, $55.2\%$ Swahili, and $58.9\%$ Yoruba. Amharic therefore enters selection with a smaller usable counterfactual pool, a point revisited in \S\ref{sec:analysis}. The validator is deliberately task-specific: extending \method{} to sequence labelling, intent/slot filling, or summarisation means defining the corresponding counterfactual transformation for that task.

\paragraph{Audit channels.}
Each candidate receives five normalised scores in $[0,1]$:
\begin{align}
U_i &= \tfrac{1}{3}(\sigma_{\text{gen}} + a_{\text{teach}} + v_{\text{nbr}}),\\
L_i &= \max\{E_i, M_i, B_i\},\\
H_i &= \mathrm{AUC}(\hat{p}_{\text{art}}),\\
D_i &= 1 - \max_{j \in G \cup K_{<i}} \cos(\mathbf{e}_i, \mathbf{e}_j),\\
C_i &= \mathbf{1}[f_\theta(x_i')=y_i'] \cdot s_{\text{cf}}.
\end{align}
The \emph{utility} score $U_i$ averages generator confidence, the agreement of a frozen XLM-R teacher \citep{conneau2020xlmr} fine-tuned on $G_{t,\ell,b}$, and a $k=5$ validation-neighbour vote in LaBSE embedding space \citep{feng2022labse}. The \emph{leakage} score $L_i$ takes the maximum of exact 10-gram overlap $E_i$, a MinHash near-duplicate score $M_i$ \citep{broder1997syntactic} at band threshold $0.85$, and LaBSE cosine similarity $B_i$ against the dev/test union. The \emph{shortcut} score $H_i$ is the AUC of a logistic probe that predicts $y_i$ from artifact features $\phi(x_i)$ (length, label-word lexicon, named-entity inventory, template indicators). \emph{Diversity} $D_i$ is one minus the maximum LaBSE cosine of $x_i$ against $G$ and the candidates already kept, $K_{<i}$. \emph{Counterfactual consistency} $C_i$ multiplies an indicator that the teacher's prediction on $x_i'$ matches $y_i'$ by a generator-side semantic-edit score $s_{\text{cf}}\in[0,1]$.

The same audit record also stores the two LLM-as-judge fields used by quality-only selectors: \emph{judged label correctness} (does the judge accept $y_i$ for $x_i$?) and \emph{judged quality} (a Likert-style rubric on fluency, factuality, and task fit). These are not part of $S_i$ below; they are reported so that judge-facing quality and downstream utility can be compared on the same selected pools.

Aggregation is defined at the selector-output level. Every audit metric reported in \S\ref{sec:results} is averaged over the selector's 64-example \emph{retained} pool, so audit channels and downstream Macro-F1 are evaluated on the same training set. Hard rejection is part of this audit protocol: a candidate is hard-rejected when it violates a fixed leakage, shortcut, or counterfactual-consistency threshold. The reported hard-reject rate is therefore the \emph{residual} threshold-violation rate inside the retained pool, a diagnostic of the selector's final output rather than the pre-filter reject rate of the original candidate pool.

\paragraph{Selectors built on the audit.}
Given the per-candidate audit record, the selectors compared in \S\ref{sec:results} are:
\begin{itemize}\itemsep0pt
  \item \textbf{Naive}: keep the first $k$ examples the generator returned (no audit use).
  \item \textbf{AlpaGasus-style}: top-$k$ by judged-quality score.
  \item \textbf{DEITA-style}: top-$k$ by a weighted combination of judged quality, complexity, and embedding diversity.
  \item \textbf{\method{}-EQ}: the original equal-budget \method{} selector (hard-reject then random fill).
  \item \textbf{\method{}-V2}: hard-reject then top-$k$ by a budgeted reranker (below).
\end{itemize}

\paragraph{Hard rejection and \method{}-V2 reranker.}
\method{}-V2 first applies hard rejection,
\[
\mathrm{reject}(x_i) = \mathbf{1}[L_i{>}\tau_L \lor C_i{<}\tau_C \lor H_i{>}\tau_H],
\]
with defaults $\tau_L{=}0.15$, $\tau_C{=}0.60$, $\tau_H{=}0.70$ fixed before the replay. Surviving candidates are ranked by
\[
S_i = \alpha_U U_i + \alpha_C C_i + \alpha_D D_i - \alpha_L L_i - \alpha_H H_i,
\]
with $\alpha_U{=}1.0$, $\alpha_C{=}0.75$, $\alpha_D{=}0.50$, $\alpha_L{=}1.0$, $\alpha_H{=}0.75$. In each cell the top-$k$ by $S_i$ is retained subject to (i) class-label balance within $\pm 10\%$ of the gold prior, (ii) at most three retained candidates per source seed, and (iii) the synthetic budget. The same constraints (i) to (iii) apply to AlpaGasus and DEITA so that retained-set sizes match exactly.

Two hyperparameter groups are calibrated on a held-out Setswana split (not part of the 8-cell replay): $\tau_L{=}0.15$ is the LaBSE-cosine value at which $99\%$ of Setswana dev sentences fall below the cutoff, and the score weights $(\alpha_U,\alpha_C,\alpha_D,\alpha_L,\alpha_H)$ equalise per-channel contributions on the same split when scores are scaled into $[0,1]$. The other two are fixed by protocol: $\tau_H{=}0.70$ follows the conventional shortcut-probe ``probe is reading the label'' AUC bar, and $\tau_C{=}0.60$ matches the threshold our generator already uses for $s_{\text{cf}}\in[0,1]$. All four groups are identical across the eight reported cells and remain fixed throughout the replay.

\paragraph{Audit trace.}
Every candidate is traceable from a paper table cell, through the claim ledger, back to its row in the per-cell replay table and the underlying audit record (\S\ref{sec:reproducibility}). This section defines the instrumentation; \S\ref{sec:results} tests how audit-ranked pools behave under downstream training.

\section{Experimental Setup}
\label{sec:experiments}

\paragraph{Tasks and languages.}
We evaluate an 8-cell replay over two classification tasks: MasakhaNEWS news topic classification \citep{adelani2023masakhanews} and AfriSenti sentiment classification \citep{muhammad2023afrisenti}, in four African languages: Amharic, Hausa, Swahili, and Yoruba. Macro-F1 is the downstream metric in every cell. Appendix Table~\ref{tab:datasets} gives benchmark details, including splits and provenance. Each task-language cell is evaluated independently; we report aggregates only after inspecting the per-cell outcomes.

\paragraph{Low-resource protocol.}
Every cell uses $b{=}64$ gold examples and a $3{\times}$ synthetic budget, so each selector ends with a retained synthetic set of 64 examples. We use a deterministic replay at seed 13 to make every selector choose from the same trace-backed candidate pool in each cell. This design turns the comparison into a within-pool test of selection criteria rather than a comparison of independently sampled generations. We re-check the headline finding with cell-level bootstrap resampling and leave-one-cell-out analysis in \S\ref{sec:robustness}.

\paragraph{Generation and scoring.}
All five selectors operate on a single shared candidate pool per cell. We score candidates as described in \S\ref{sec:method}: generator and judge quality signals, XLM-R teacher agreement \citep{conneau2020xlmr}, LaBSE-based diversity and leakage checks \citep{feng2022labse}, MinHash near-duplicate checks \citep{broder1997syntactic}, shortcut features, and counterfactual-consistency scores.

\paragraph{Selectors.}
The five selectors read from the same candidate pool under identical class-balance and source-seed constraints:
\textbf{(1) Naive synthetic} keeps the first 64 synthetic candidates without audit-aware ranking;
\textbf{(2) AlpaGasus-style} \citep{chen2024alpagasus} keeps the top-64 by judged quality;
\textbf{(3) DEITA-style} \citep{liu2024deita} keeps the top-64 by quality, complexity, and diversity;
\textbf{(4) \method{}-EQ} runs the original equal-budget \method{} selector (hard-reject then random fill);
\textbf{(5) \method{}-V2} runs the budget-preserving reranker (hard-reject then top-64 by $S_i$).
A Gold-only reference fine-tuned on the 64 gold examples alone is reported in Appendix~\ref{app:cosda_details}; it is excluded from the main-body comparison because, with no synthetic data to audit, it cannot participate in the audit-vs-downstream rank correlation.

\paragraph{Models and metrics.}
For downstream evaluation, we fine-tune the deterministic Hugging Face classification path used in the \method{} replay and report Macro-F1. In every cell the classifier is fine-tuned on the union of the 64 gold seeds and the selector's 64 retained synthetic examples (128 training rows total); the gold seeds are identical across selectors, so the only thing that varies is the retained synthetic set. We compute selection-quality metrics over each selector's retained synthetic examples: judged label correctness, judged quality, counterfactual consistency $C$, leakage $L$, shortcut $H$, and hard-reject rate.

\paragraph{Ground-truth artifacts.}
We compute every numeric claim in the paper from the full per-cell replay table (selectors $\times$ task-language cells, with audit channels and downstream Macro-F1). A released claim ledger maps each table cell and in-text aggregate to its source row. Appendix~\ref{app:reproducibility} lists the ground-truth artifacts.

\section{Results: The Quality-to-Utility Gap}
\label{sec:results}

Two judge quantities drive this section, both read off each selector's retained pool: \emph{label correctness} (LC), the fraction of retained examples whose proposed label the judge accepts, and \emph{quality} ($Q$), the judge's Likert rating of fluency, factuality, and task fit. The result is one empirical claim: ranking the five selectors by these audit scores and by downstream Macro-F1 yields two orderings that disagree, and the disagreement survives every robustness check we run in this setting. \S\ref{sec:gap} presents the aggregate mismatch; \S\ref{sec:headtohead} puts the label-correctness leader (\method{}-V2) head-to-head with the downstream leader (AlpaGasus); \S\ref{sec:robustness} confirms the pattern at the cell level.

\subsection{The gap}
\label{sec:gap}

\begin{table}[t]
\centering
\small
\begin{tabular}{lrrrr}
\toprule
\textbf{Baseline} & \textbf{LC$\uparrow$} & \textbf{Q$\uparrow$} & \textbf{Hard Rej.$\downarrow$} & \textbf{F1$\uparrow$} \\
\midrule
Naive       & 0.767 & 0.662 & 0.486 & 0.167 \\
AlpaGasus   & 0.890 & 0.749 & 0.246 & \textbf{0.202} \\
DEITA       & 0.869 & 0.737 & 0.299 & 0.128 \\
\method{}-EQ   & 0.836 & 0.718 & \textbf{0.109} & 0.133 \\
\method{}-V2   & \textbf{0.904} & 0.746 & 0.162 & 0.163 \\
\midrule
\multicolumn{4}{l}{\textit{Spearman} $\rho$ \textit{(LC-rank vs.\ F1-rank)}} & $+0.10$ \\
\bottomrule
\end{tabular}
\caption{Audit and downstream summary across the five selectors, averaged over 8 task-language cells. LC = judged label correctness; Q = judged quality; F1 = mean Macro-F1. \textbf{Bold} marks the best selector in each column.}
\label{tab:gap_summary}
\end{table}

Table~\ref{tab:gap_summary} is the headline result. On the three summary audit channels, every quality-aware selector improves over naive: judged label correctness and judged quality rise, and the hard-reject rate falls. Here and throughout, the reported hard-reject rate is the \emph{residual} rate inside the already-retained 64-example pool, i.e.\ the fraction of retained examples that still violate a leakage, shortcut, or counterfactual-consistency threshold; it is a diagnostic of the selector's final output, not a pre-filter rejection rate on the raw candidate pool (\S\ref{sec:method}). Macro-F1 imposes a different order. AlpaGasus has the highest downstream mean ($+0.036$ over naive), while DEITA, \method{}-EQ, and \method{}-V2 fall below naive despite better audit profiles. Concretely, the LC order is V2 $>$ AlpaGasus $>$ DEITA $>$ EQ $>$ naive, whereas the Macro-F1 order is AlpaGasus $>$ naive $>$ V2 $>$ EQ $>$ DEITA. The selector-level Spearman between mean LC and mean Macro-F1 is $\rho{=}0.10$, and the same correlation between mean hard-reject rate and mean Macro-F1 is $\rho{=}0.20$; because these are computed over only five selectors, we read them qualitatively (as the rank orders above) and treat the within-cell Spearman below as the primary inferential statistic. The disagreement extends beyond the judge: Spearman between $C$ (counterfactual consistency) and Macro-F1 is $\rho{=}-0.20$, and the same correlation for a composite of the three non-judge channels ($C$, $H$, hard-reject) against Macro-F1 is also $\rho{=}-0.20$. The audit and the downstream task are measuring different properties of the selected pool.

\begin{table}[t]
\centering
\footnotesize
\setlength{\tabcolsep}{3.5pt}
\begin{tabular}{lrrrrr}
\toprule
\textbf{Selector} & \textbf{LC$\uparrow$} & \textbf{Q$\uparrow$} & $\mathbf{C}\uparrow$ & $\mathbf{H}\downarrow$ & \textbf{Rej.$\downarrow$} \\
\midrule
Naive       & 0.767 & 0.662 & 0.587 & 0.066 & 0.486 \\
AlpaGasus   & 0.890 & \textbf{0.749} & 0.653 & 0.065 & 0.246 \\
DEITA       & 0.869 & 0.737 & 0.637 & 0.070 & 0.299 \\
\method{}-EQ   & 0.836 & 0.718 & \textbf{0.686} & 0.058 & \textbf{0.109} \\
\method{}-V2   & \textbf{0.904} & 0.746 & 0.674 & \textbf{0.057} & 0.162 \\
\bottomrule
\end{tabular}
\caption{Audit-quality channels for selected synthetic data, averaged over 8 cells. LC = judged label correctness; Q = judged quality; $C$ = counterfactual consistency; $H$ = shortcut score; Rej. = hard-reject rate. \textbf{Bold} marks the best selector in each column.}
\label{tab:audit_quality_main}
\end{table}

Table~\ref{tab:audit_quality_main} gives the broader audit-channel picture: \method{}-V2 leads on LC and shortcut $H$, \method{}-EQ leads on $C$ and hard-reject, and AlpaGasus is marginally best on $Q$. The four quality-aware selectors all separate from naive on LC, $Q$, $C$, and hard-reject, and the shortcut scores further separate the \method{} variants from quality-only ranking. Only AlpaGasus also converts its audit profile into a downstream gain over naive. The pattern is ``better audit, different retained pool,'' rather than ``better audit, better model.'' These numbers come from one controlled configuration --- eight cells at a single seed, one generator, one judge, one backbone (XLM-RoBERTa base), and no human validation of the synthetic text --- with low, sometimes near-degenerate Macro-F1, so we read the tables for rank \emph{disagreement} rather than for a selector-vs-naive win (\S\ref{sec:limitations} gives the full scope).

\subsection{Head-to-head: LC leader vs.\ downstream leader}
\label{sec:headtohead}

\begin{table}[t]
\centering
\footnotesize
\setlength{\tabcolsep}{4pt}
\begin{tabular}{lrrr}
\toprule
\textbf{Attribute} & \textbf{AlpaGasus} & \textbf{\method{}-V2} & $\boldsymbol{\Delta}$ \\
\midrule
Macro-F1$\uparrow$         & \textbf{0.202} & 0.163 & $+0.039$ \\
LC$\uparrow$               & 0.890 & \textbf{0.904} & $-0.014$ \\
Q$\uparrow$                & \textbf{0.749} & 0.746 & $+0.003$ \\
Hard reject$\downarrow$      & 0.246 & \textbf{0.162} & $+0.084$ \\
$C{\uparrow}$              & 0.653 & \textbf{0.674} & $-0.021$ \\
Per-cell wins$\uparrow$    & 3 & 3 & (2 ties) \\
\bottomrule
\end{tabular}
\caption{Head-to-head between the LC leader (\method{}-V2) and the downstream leader (AlpaGasus). \textbf{Bold} marks the better value per row.}
\label{tab:head_to_head}
\end{table}

Table~\ref{tab:head_to_head} isolates the core inversion. \method{}-V2 leads AlpaGasus on LC, hard-reject, and $C$, and trails by only $0.003$ on $Q$; AlpaGasus nevertheless carries a $+0.039$ mean Macro-F1 lead. Per-cell, the two selectors tie 3-3-2, so the downstream edge is driven by effect size rather than broad dominance. The audit identifies a cleaner pool, while Macro-F1 rewards a different selected set.

\begin{figure}[t]
\centering
\includegraphics[width=\linewidth]{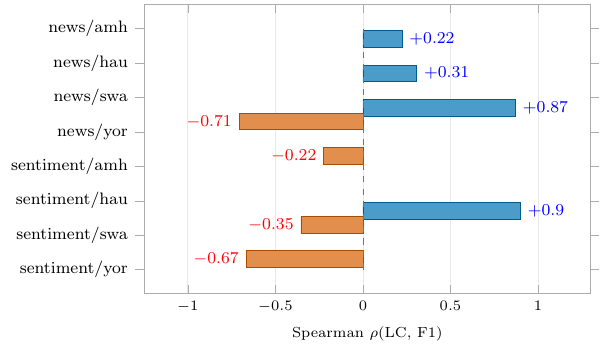}
\caption{Within-cell Spearman rank correlation between judged label correctness and downstream Macro-F1, computed across the five selectors within each of the eight task-language cells.}
\label{fig:within_cell_spearman}
\end{figure}

\paragraph{Within-cell ranks.}
To rule out aggregation effects, we also compute the within-cell Spearman between LC and Macro-F1 inside each of the eight cells (Figure~\ref{fig:within_cell_spearman}). Four cells are positive (Swahili news $+0.87$, Hausa sentiment $+0.90$, news/amh, news/hau) and four are negative (Yoruba news, Amharic sentiment, Swahili sentiment, Yoruba sentiment); the mean is $+0.04$ and the median is $0.00$. The disagreement is present inside individual cells, not only in the aggregate.

\subsection{Robustness}
\label{sec:robustness}

\begin{figure}[t]
\centering
\includegraphics[width=\linewidth]{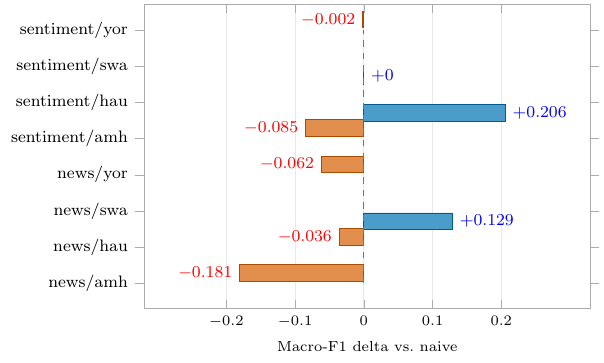}
\caption{Per-cell Macro-F1 delta of \method{}-V2 relative to naive. Each bar is one task-language cell; positive values mean \method{}-V2 beats naive.}
\label{fig:per_cell_delta}
\end{figure}

\begin{table}[t]
\centering
\footnotesize
\setlength{\tabcolsep}{4pt}
\begin{tabular}{lrrrr}
\toprule
\textbf{Cell} & \textbf{Naive$\uparrow$} & \textbf{AlpaGasus$\uparrow$} & \textbf{V2$\uparrow$} & \textbf{V2$-$Naive$\uparrow$} \\
\midrule
news/amh        & 0.286 & 0.342 & 0.105 & $-0.181$ \\
news/hau        & 0.102 & 0.195 & 0.066 & $-0.036$ \\
news/swa        & 0.069 & 0.184 & 0.198 & $+0.129$ \\
news/yor        & 0.128 & 0.065 & 0.065 & $-0.062$ \\
sent.\,/amh     & 0.154 & 0.067 & 0.069 & $-0.085$ \\
sent.\,/hau     & 0.168 & 0.339 & 0.375 & $+0.206$ \\
sent.\,/swa     & 0.248 & 0.248 & 0.248 & $+0.000$ \\
sent.\,/yor     & 0.178 & 0.176 & 0.176 & $-0.002$ \\
\bottomrule
\end{tabular}
\caption{Per-cell Macro-F1 for naive, AlpaGasus, and \method{}-V2 in the full replay; DEITA, \method{}-EQ, and the Gold-only reference are in Appendix~\ref{app:cosda_details}.}
\label{tab:per_cell}
\end{table}

\begin{table}[t]
\centering
\small
\begin{tabular}{lrr}
\toprule
\textbf{Cell removed} & \textbf{Mean $\Delta$ (V2$-$Naive)$\uparrow$} \\
\midrule
\textit{(all 8 cells)} & $-0.004$ \\
news/amh        & $+0.022$ \\
news/hau        & $+0.001$ \\
news/swa        & $-0.023$ \\
news/yor        & $+0.005$ \\
sentiment/amh   & $+0.008$ \\
\textbf{sentiment/hau}   & $\mathbf{-0.034}$ \\
sentiment/swa   & $-0.004$ \\
sentiment/yor   & $-0.004$ \\
\bottomrule
\end{tabular}
\caption{Leave-one-cell-out mean $\Delta$ of \method{}-V2 vs.\ naive. The first row is the all-cell baseline; subsequent rows show the mean when the named cell is excluded.}
\label{tab:loo_delta}
\end{table}

\paragraph{Per-cell heterogeneity.}
Figure~\ref{fig:per_cell_delta} shows that the aggregate $-0.004$ mean is structured: two cells are positive (Swahili news $+0.129$, Hausa sentiment $+0.206$) and six are negative or near-zero. Table~\ref{tab:per_cell} gives the underlying per-cell Macro-F1; three cells (news/yor, sent./swa, sent./yor) are near-degenerate, with five of six baselines collapsing to the same value. The audit-utility gap appears at the cell level, not as a uniform near-miss against naive.

\paragraph{Bootstrap CI.}
Resampling cells ($B{=}10{,}000$) gives 95\% confidence intervals on the mean delta over naive of $[-0.079,+0.078]$ for \method{}-V2 and $[-0.020,+0.093]$ for AlpaGasus. On the five non-degenerate cells alone the intervals are $[-0.114,+0.129]$ for V2 (mean $+0.007$) and $[-0.015,+0.137]$ for AlpaGasus (mean $+0.070$). The intervals place the selector-vs-naive mean differences at cell-level uncertainty. The audit-vs-downstream disagreement is the directly testable signal in this design, and it is the signal that holds.

\paragraph{Leave-one-cell-out.}
The eight-cell V2$-$naive mean of $-0.004$ sits near zero because Hausa sentiment contributes a large positive effect ($+0.206$); removing that cell moves the mean to $-0.034$ (Table~\ref{tab:loo_delta}), comparable to the negative aggregate for DEITA and \method{}-EQ. The all-cell average is built from cell-level effects of opposite sign rather than a uniform selector advantage, exactly as Figure~\ref{fig:per_cell_delta} shows.

\paragraph{Non-degenerate cells.}
Dropping the three cells where the classifier collapses (Yoruba news, Swahili sentiment, Yoruba sentiment) leaves five informative cells. On these, mean F1 is $0.226$ (AlpaGasus), $0.163$ (V2), $0.156$ (naive), $0.114$ (EQ), $0.103$ (DEITA), with the Gold-only reference at $0.087$. The LC-to-F1 Spearman climbs to $+0.50$, and the AlpaGasus-vs-V2 inversion holds with stronger margin. The gap finding strengthens, rather than weakens, on the informative subset.

\paragraph{Audit improvements.}
The audit-channel gains over naive in Table~\ref{tab:audit_quality_main} are clean properties of each selected pool (\method{}-V2 lifts LC by $+17.9\%$ relative; \method{}-EQ cuts hard-reject by $77.6\%$ relative and raises $C$ from $0.587$ to $0.686$); \S\ref{sec:analysis} explains how they coexist with a different downstream order. The leakage column and per-cell DEITA/\method{}-EQ values are in Appendix~\ref{app:cosda_details}.

\section{Analysis: Why the Audit Misses}
\label{sec:analysis}

A selector can lead on every key audit channel and still finish mid-pack downstream because the audit and the downstream task measure different properties of the selected pool. We point to three mechanisms visible in the trace data, each supported by the audit channels themselves.

\paragraph{H1: Audit-strict selectors prune away support the classifier needs.}
\method{}-V2 has the lowest shortcut score and the second-lowest hard-reject rate (behind \method{}-EQ). The candidates it removes are, by construction, the audit's worst, but they are not always the downstream classifier's worst. In the five cells where V2 loses to naive, the residual hard-reject rate inside V2's own pool averages $0.23$; in the two cells where V2 wins, it averages $0.08$. Treating cell-level hard-reject rate as a covariate gives Pearson correlation $-0.56$ with the V2$-$naive Macro-F1 delta (two-sided permutation $p{=}0.15$, $N{=}100{,}000$). For an eight-cell trace, the useful signal is the direction and alignment with the per-cell cases: higher residual reject inside V2's pool tracks worse V2-vs-naive downstream. The controlled ablation is direct: graft V2's hard-reject onto AlpaGasus's ranking.

\paragraph{H2: AlpaGasus does not win on variety.}
A natural explanation for AlpaGasus's downstream lead is that quality ranking preserves more topical or lexical variety than the \method{} selectors. The diversity column in Appendix Table~\ref{tab:audit_quality_full} rejects that explanation: AlpaGasus's $D{=}0.590$ is the lowest of the five selectors, below naive ($0.593$) and every \method{} variant ($D{\in}[0.603,0.604]$); DEITA, which explicitly weights diversity, is highest at $0.649$. The audit's own diversity channel therefore rules out variety preservation as the mechanism behind AlpaGasus's downstream advantage. The remaining signal is topical or decision-boundary support in candidates that score high on judged quality but are filtered out by V2's counterfactual or shortcut gates, precisely the region where audit strictness and training utility diverge.

\paragraph{H3: LC saturates and loses resolution.}
Judged LC is $\geq 0.77$ for every selector and saturates around $0.90$; judged $Q$ is compressed into the range from $0.66$ to $0.75$. Both channels rank selectors confidently, with rank correlation between LC and $Q$ across selectors at $+0.90$, but at this compression they cannot separate selectors at the resolution downstream Macro-F1 demands. The audit-utility gap is therefore visible on the judge channels alone, and the non-judge channels ($C$, $H$, hard-reject) reproduce the same disagreement (\S\ref{sec:gap}). The finding is not confined to judge-derived scores.

\paragraph{Putting H1 to H3 together.}
The three mechanisms compose: V2 prunes harder than naive (H1), AlpaGasus wins through quality-correlated topical signal that V2 filters out rather than through variety preservation (H2), and LC and $Q$ saturate at the top (H3). The audit is therefore a complementary instrument for exposing selected-pool properties, not a substitute for downstream validation.

\subsection{Cell-level case studies}
\label{sec:case_studies}

The eight cells split into three groups. \textbf{Positive (2):} Swahili news ($+0.129$) and Hausa sentiment ($+0.206$), where naive Macro-F1 is moderate and \method{}-V2's residual hard-reject rate is low ($0.16$, $0.00$), so V2 adds signal without sacrificing support. \textbf{Negative (3):} Amharic news ($-0.181$), Hausa news ($-0.036$), and Amharic sentiment ($-0.085$), where V2's audit beats naive but Macro-F1 is worse; the two largest losses coincide with the largest residual hard-reject rates inside V2's pool ($0.609$, $0.406$), and Amharic has the lowest $C_i$ pass rate ($33.4\%$ vs.\ $52\%$--$59\%$ elsewhere, \S\ref{sec:method}), so V2's hard-reject strips a larger fraction of its pool before ranking --- the clearest evidence for H1. \textbf{Near-degenerate (3):} Yoruba news, Swahili sentiment, Yoruba sentiment, where five of six baselines collapse to identical Macro-F1 and selector differences carry little information (\S\ref{sec:robustness} shows the gap finding strengthens when these are dropped). The groups line up with the naive Macro-F1 scale, so a benchmark in this regime should separate cells that exercise the audit from cells that exercise the classifier.

\subsection{What the audit pairs with}
\label{sec:prescription}

Our replay supports two direct recommendations:
\begin{enumerate}\itemsep0pt
  \item \textbf{Report both audit and downstream metrics on the same selected sets.} Quality-only papers rarely expose this comparison, and an audit pipeline makes it straightforward.
  \item \textbf{Treat LLM-as-judge label correctness as an audit channel, not a downstream proxy.} The selector-level Spearman in this replay is $\rho{=}0.10$. The audit is valuable because it reveals selected-pool properties directly.
\end{enumerate}
Three follow-up directions for closing the gap are sketched in \S\ref{sec:conclusion}: layering training-set coverage objectives on the audit, reranking against downstream-validation behaviour, and joint optimisation over audit and downstream signals.

\paragraph{When to prefer \method{}-V2.}
V2 is the right selector when audit channels carry compliance weight: when leakage is a hard constraint (V2's $L{=}0.003$ vs.\ AlpaGasus's $0.006$), when shortcut robustness matters at inference (V2's $H{=}0.057$ is lowest), or when label correctness is a public-facing claim (V2's LC$=0.904$ is the only entry above $0.90$). These are real selected-pool properties, even when a different selector maximises mean Macro-F1.

\paragraph{Filtering strictness vs.\ diversity.}
The two \method{} variants vary filtering strictness on shared channels: \method{}-EQ hard-rejects then random-fills, V2 hard-rejects then ranks by $S_i$. They reach similar $D$ ($0.604$ vs.\ ${\approx}0.60$) but different hard-reject rates ($0.109$ vs.\ $0.162$), with downstream means inside each other's bootstrap uncertainty. The channel that moves Macro-F1 over naive is the quality-only ranking step itself, used by AlpaGasus without hard-reject. The clean follow-up is to graft V2's hard-reject onto AlpaGasus's ranking; the released traces set up that test.

\section{Related Work}
\label{sec:related_work}

\paragraph{Synthetic supervision and selection.}
Language models can generate supervision from seed examples or task descriptions \citep{wang2023selfinstruct,honovich2023unnatural,wang2022supernatural,mishra2022cross,bach2022promptsource,longpre2023flan,taori2023alpaca,xu2023wizardlm}, and multilingual variants extend this recipe across languages \citep{ustun2024aya,singh2024aya23}. A second line asks which generated examples to keep: AlpaGasus ranks by GPT-4 quality \citep{chen2024alpagasus}; DEITA combines quality, complexity, and diversity \citep{liu2024deita}; LESS selects by validation influence \citep{xia2024less}; QuRating learns a quality model \citep{wettig2024qurating}; and related work studies self-guided scoring and small curated sets \citep{li2024quantity,cao2024instruction,zhou2023lima}. These methods largely inherit the same assumption: selector quality predicts downstream utility. Our replay tests that assumption under matched budget in low-resource multilingual classification.

\paragraph{LLM judges as selection evidence.}
LLM judges are attractive for synthetic-data selection because they scale cheaply across candidate pools and return interpretable quality signals. The critical distinction is what the score certifies: high label correctness means the judge can map a candidate to the intended label, not that a small downstream classifier will learn a better decision boundary from that candidate. This distinction is sharper in our setting because each retained pool has 64 synthetic examples. One mislabeled, duplicated, or shortcut-heavy cluster can matter, and so can a noisy example that expands topical coverage. We therefore treat judge scores as audit channels rather than final evidence of training utility.

\paragraph{Utility-vs-quality misalignment.}
Adjacent work has flagged the same tension: DEITA combines quality with complexity and diversity because quality alone is insufficient \citep{liu2024deita}, LESS side-steps the issue by selecting against validation influence \citep{xia2024less}, and data-centric studies document cases where ``cleaner'' subsets underperform noisier supersets \citep{swayamdipta2020dataset,sorscher2022beyond}. We make the misalignment explicit at the \emph{selector} level, in a low-resource multilingual setting where the cost of a noisy retained example is amplified, and quantify it via within-cell rank statistics on identical candidate pools.

\paragraph{Data-centric diagnostics and counterfactual checks.}
Data-centric NLP shows that example choice can matter as much as model choice, using cartography, forgetting, proxy selection, learnability, and pruning to find useful or harmful points \citep{swayamdipta2020dataset,toneva2019forgetting,coleman2019selection,paul2021deep,mindermann2022prioritized,sorscher2022beyond}. Counterfactual and behavioral tests ask whether models respond to task-relevant changes rather than artifacts \citep{kaushik2020learning,gardner2020evaluating,ribeiro2020checklist,wu2021polyjuice}, while work on shortcuts, contamination, and memorisation shows why fluent examples can still be risky \citep{gururangan2018annotation,poliak2018hypothesis,mccoy2019right,geirhos2020shortcut,jia2017adversarial,dodge2021documenting,carlini2021extracting,lee2022deduplicating,kandpal2022deduplicating,maini2024rephrasing}. We move these diagnostics from model evaluation to synthetic-data selection: each retained pool is audited for leakage, shortcuts, and counterfactual consistency before we ask whether those audit signals predict Macro-F1.

\paragraph{Low-resource multilingual benchmarks.}
The replay uses MasakhaNEWS for African news classification and AfriSenti for African-language sentiment \citep{adelani2023masakhanews,muhammad2023afrisenti}. The four covered languages span two families (Afro-Asiatic, Niger-Congo) and two scripts (Ge'ez, Latin), so the replay exercises selector behaviour across script and morphology rather than topic alone. Extending the audit beyond these classification cells calls for task-specific counterfactual validators for sequence labelling, intent/slot filling, or summarisation.

\section{Conclusion}
\label{sec:conclusion}

This replay separates two properties that synthetic-data selection often conflates: audit quality and downstream utility. \method{}-V2 leads judged label correctness yet trails downstream, while AlpaGasus inverts that order; bootstrap, leave-one-cell-out, and non-degenerate-cell checks preserve the gap. The contribution is diagnostic: in this controlled setting, synthetic-data evaluations should report audit and downstream metrics on the same retained sets, and the next selector family should pair audit constraints with coverage or validation-set utility. We conjecture, but do not test here, that the same decoupling can arise wherever a quality proxy is scored independently of the downstream model --- filtered pretraining, instruction-tuning curation, and preference-data pipelines share that structure --- and the released traces are intended to support that broader replay.

\pagebreak
\section*{Limitations}
\label{sec:limitations}

\paragraph{Cell count and seed.}
The replay covers 8 task-language cells from MasakhaNEWS and AfriSenti at a single seed. The selector-level Spearman $\rho{=}0.10$ is computed over 5 selectors, and the bootstrap CIs on V2$-$naive and AlpaGasus$-$naive both contain zero. These intervals keep the paper's downstream claim focused on rank disagreement rather than a selector-vs-naive win. The deterministic Hugging Face replay across $5{\times}8{=}40$ fine-tunes plus the LLM-judge audit costs roughly one GPU-week per seed; we use that budget to keep selectors tied to identical candidate pools and to run cell-level robustness checks. The within-cell Spearman is computed independently in each cell, so it is robust to between-cell variation; this is the inferential statistic on which the gap finding rests.

\paragraph{Task scope.}
The completed replay covers two classification tasks. Sequence labelling, intent/slot filling, and summarisation require task-specific counterfactual validators (\S\ref{sec:method}); the present study isolates the classification setting before extending the validator family.

\paragraph{Judge and prompt dependence.}
Judged label correctness and judged quality come from a single multilingual LLM-as-judge prompted in English with in-language task descriptions. The audit-table numbers we report (LC saturating near $0.90$, $Q$ from $0.66$ to $0.75$) are properties of that judge and prompt, so a multi-judge or multi-prompt sweep could change the LC ordering. The gap finding does not depend on that channel alone. First, the disagreement is also present on non-judge channels (hard-reject, $C$, $L$, $H$), which are computed without the judge: selector-level Spearman against Macro-F1 is $\rho{=}0.20$ for hard-reject and $\rho{=}-0.20$ for $C$ and for the non-judge composite. Second, the within-cell Spearman is computed inside each cell on the same candidate pool, so judge-induced shifts in LC level do not drive the rank-correlation result. Multi-judge auditing is the natural robustness extension for deployment settings that rely on LC as a public-facing quality claim.

\paragraph{Leakage and shortcut detectors.}
Exact $n$-gram, MinHash, and LaBSE-based leakage checks target observable contamination; they do not certify the absence of memorisation in upstream generator training data. The shortcut probe uses a fixed feature inventory. All five selectors read the same audit fields, so detector imperfections affect the interpretation of audit-quality differences across selectors uniformly.

\paragraph{Generator and backbone.}
All synthetic candidates come from Qwen2.5-14B-Instruct with one nucleus-sampling configuration, and the downstream classifier is XLM-RoBERTa base \citep{conneau2020xlmr}. The headline finding is therefore scoped to this generator/backbone pair. A stronger generator or backbone could re-align the rankings; the audit instrumentation is designed to test that transfer directly.

\paragraph{Score-weight calibration.}
We chose the score weights $(\alpha_U,\alpha_C,\alpha_D,\alpha_L,\alpha_H)$ in $S_i$ on a held-out Setswana development split to equalise per-channel contributions in $[0,1]$. We ship them with the claim ledger, so subsequent analyses can vary them without regenerating candidates. A $\pm10\%$ sensitivity sweep on $\tau$ and $\alpha$ would quantify how much the V2 retained pool moves under nearby protocol choices; aggregating the per-candidate $D_i$ into Appendix Table~\ref{tab:audit_quality_full} already closes one previously deferred channel check.

\paragraph{Baseline coverage.}
The compared selectors are matched-budget reimplementations of each method's \emph{selection criterion} (AlpaGasus's judged-quality ranking, DEITA's quality/complexity/diversity mix), not runs of the original codebases; all read the shared audit record so the comparison isolates the selection rule. We do not yet include a validation-aware or influence-based selector such as LESS \citep{xia2024less}. Because our headline concerns downstream utility, a validation-aware selector is the most relevant missing comparison; the released audit records and selection harness are set up to add it directly, and it is the priority extension.

\paragraph{Causal claims.}
The H1 to H3 mechanisms in \S\ref{sec:analysis} are trace-grounded explanations, not causal ablations. A controlled follow-up such as AlpaGasus$+$\method{}-V2 hard-reject would isolate which audit step drives which downstream effect, and the released audit records make that intervention straightforward.

\section*{Ethical Considerations}

Synthetic data for low-resource communities can ease annotation pressure, but it can also distort cultural context, normalise incorrect language use, or surface private information from web-trained generators \citep{bender2021dangers}. \method{} addresses these risks as an audit instrument: it logs provenance, checks leakage at three levels, and measures counterfactual consistency before an example enters the retained pool.

Dataset licenses are recorded in the released manifest. The replay uses MasakhaNEWS and AfriSenti, and generated examples are marked as generated in the released artifacts. A data statement \citep{bender2018datasheets} and a datasheet \citep{gebru2021datasheets} accompany the released audit logs, listing language coverage, intended use, known failure modes, and a removal procedure for individuals who identify content they did not consent to release.

This paper reports automatic audit metrics rather than a human annotation study, so it does not claim human-validated cultural plausibility. Deployment of selected examples should include community review by speakers of the target language, especially for political news and sentiment data.

\appendix
\begingroup
\emergencystretch=3em
\section{Reproducibility Checklist}
\label{sec:reproducibility}
\label{app:reproducibility}

\paragraph{How to read the appendix.}
The appendix is organised as a trace from evidence to claim. This section names the source artifacts and replay contract; Appendix~\ref{app:annotation} describes the prompts and audit fields that create the per-candidate records; Appendix~\ref{app:cosda_details} gives the full detail tables and appendix plot. The main paper reports the claim-carrying summaries; the appendix exposes the rows and artifacts behind them.

\paragraph{Ground-truth replay.}
Every numeric claim in the paper is computed from one replay artifact: a per-cell table indexed by task, language, and selector. Each row records the retained-set size, judged label correctness, judged quality, counterfactual consistency $C$, leakage score $L$, shortcut score $H$, hard-reject rate, and downstream Macro-F1 from the deterministic Hugging Face classification path. The same table populates the main-body tables, appendix detail tables, and appendix figures. Aggregates (selector-level means, Spearman correlations, bootstrap CI, leave-one-cell-out, and non-degenerate-subset statistics) are deterministic functions of this table, not separate experiments.

\paragraph{Ground-truth artifact paths.}
Table~\ref{tab:artifact_sources} lists the files treated as authoritative for the paper. The top-level simulated \path{results/*.json} files are not used as evidence in this revision.

\begin{table*}[t]
\centering
\small
\begin{tabular}{p{0.42\linewidth}p{0.48\linewidth}}
\toprule
\textbf{Artifact} & \textbf{Use in the paper} \\
\midrule
\path{CoSDA/cosda_emnlp_paper_ready_summary_20260521.md} &
Human-readable summary of the final CoSDA-supported claims, including the audit/downstream mismatch and scope conditions. \\
\path{CoSDA/reports/aws_collect_20260521/server_34_207_187_226/aws_vllm_revised_select1_v2_full_20260521_analysis_rows.csv} &
Authoritative row-level replay table for selectors, task-language cells, audit channels, and deterministic downstream Macro-F1. \\
\path{CoSDA/results/cosda_paper_tables_20260521.tex} &
Generated table source used to cross-check the downstream and audit summary tables before the manuscript tables were polished. \\
\bottomrule
\end{tabular}
\caption{Ground-truth artifacts used for the reported CoSDA claims. All reported numbers should be recoverable from these files or deterministic aggregations of them.}
\label{tab:artifact_sources}
\end{table*}

\paragraph{Released artifacts.}
We release four artifact groups: (i) a dataset manifest covering source, license, split, checksum, and task-language metadata for MasakhaNEWS and AfriSenti; (ii) the generated candidate examples and per-candidate audit records (utility components, leakage components, shortcut features, diversity, counterfactual consistency, judged label correctness, judged quality, and hard-reject decision); (iii) the per-cell replay table described above for all five selectors plus the gold-only reference; and (iv) the scripts that download data, generate candidates, score audit channels, apply each selector, fine-tune the downstream classifier, and emit the replay table. The release separates candidate-level audit records from selected-set summaries so that a reader can recompute a selector without trusting the aggregate tables.

\paragraph{Claim ledger.}
Every number in the abstract, results, analysis, robustness, and conclusion sections is recorded in a released claim ledger. The ledger links each table cell or in-text value to the row of the replay table, or to the aggregate statistic, from which it was computed. It provides the audit point for any subsequent re-analysis of the findings.

\paragraph{Regenerating appendix artifacts.}
The script \path{scripts/make_cosda_appendix_artifacts.py} regenerates the CoSDA-derived plot artifacts. It reads the authoritative CSV above and writes the CoSDA-labelled figure files under \path{figures/}. The older \path{scripts/make_artifacts.py} path is not part of the evidence pipeline for this paper because it targets legacy simulated artifacts.

\paragraph{Default hyperparameters.}
For the audit channels, $\alpha_U{=}1.0$, $\alpha_C{=}0.75$, $\alpha_D{=}0.50$, $\alpha_L{=}1.0$, and $\alpha_H{=}0.75$; hard-reject thresholds are $\tau_L{=}0.15$, $\tau_C{=}0.60$, and $\tau_H{=}0.70$. These values were fixed before the replay and reused unchanged across all eight task-language cells. Downstream fine-tuning uses the deterministic Hugging Face classification path with default learning rate and batch size; the seed is 13 and the gold budget is $b{=}64$ with synthetic multiplier $m{=}3$.

\paragraph{Calibration of thresholds and weights.}
The threshold-and-weight calibration is described in \S\ref{sec:method} (paragraph ``Hard rejection and \method{}-V2 reranker''). The summary: $\tau_L$ and the $\alpha$ weights are calibrated on a held-out Setswana split; $\tau_H$ and $\tau_C$ are fixed by protocol. All four groups are identical across the eight reported cells and remain fixed throughout the replay.

\paragraph{Deferred experiments.}
Multi-seed confidence intervals, human annotation, generator replacement, budget sweeps, and task extensions to NER, intent/slot filling, and summarisation are not claimed as completed results in this paper.

\section{Prompts and Audit Protocol}
\label{app:annotation}

\paragraph{Purpose.}
The prompt and audit protocol is designed to make synthetic examples inspectable before selection. The generator creates candidate examples and counterfactual edits; the audit record then stores both judge-facing quality fields and non-judge diagnostic fields. The downstream classifier never sees the audit fields directly. It sees only the retained synthetic examples chosen by each selector.

\paragraph{Generation prompt template.}
The prompt structure and the forbidden-behaviour clause are reproduced in \S\ref{sec:method} (paragraph ``Setup''). The prompt hash stored in each candidate record is used for traceability, not as a modelling feature.

\paragraph{Counterfactual edit template.}
For each candidate $x_i$, the generator receives $x_i$ and produces $x_i'$ by changing one classification-relevant variable, such as the topic or sentiment label cue, while preserving language, register, and as much non-label content as possible. The expected flipped label is stored with the counterfactual pair and used by the counterfactual-consistency score. A pair is therefore evaluated as a data-quality check: the candidate should be fluent and labelled correctly, and the paired edit should move the label in the expected direction.

\paragraph{Automatic audit fields.}
The replay records judged label correctness, judged quality, counterfactual consistency ($C$), leakage score ($L$), shortcut score ($H$), hard-reject status, and selected-set membership for each retained-set baseline. These fields support Tables~\ref{tab:gap_summary} and~\ref{tab:audit_quality_full}. They are automatic audit signals, not human annotation outcomes.

\paragraph{Selector membership fields.}
Each selector receives the same candidate pool within a task-language cell. The audit record stores whether a candidate is retained by naive, AlpaGasus-style, DEITA-style, \method{}-EQ, or \method{}-V2 selection. This shared-pool design is why the appendix tables can compare audit quality and downstream Macro-F1 without confounding the result with different generation runs.

\paragraph{Hard-reject interpretation.}
The interpretation of the hard-reject column, an audit decision over each selector's retained pool rather than a ground-truth label, is described in \S\ref{sec:method}. The downstream results separate this retained-pool property from Macro-F1.

\paragraph{Qualitative examples.}
Table~\ref{tab:qualitative_examples} gives four candidates drawn from the per-candidate audit records, illustrating one example each from the categories \emph{kept}, \emph{rejected for counterfactual}, \emph{rejected for shortcut}, and \emph{rejected for leakage}. The candidates are shown with a brief description of the trigger, not the raw text, since text in four African languages is hard to read inline. The release ships the full audit record for every candidate, including the raw text, the generator's counterfactual edit, the per-channel scores, and the hard-reject reason.

\begin{table}[h]
\centering
\footnotesize
\setlength{\tabcolsep}{4pt}
\begin{tabular}{p{0.18\linewidth}p{0.74\linewidth}}
\toprule
\textbf{Status} & \textbf{Trigger and audit profile} \\
\midrule
\textit{Kept} & Hausa business candidate; LC$=0.8$, $C{=}0.74$ (teacher predicts the flipped label on the counterfactual), $L{=}0.00$, $H{=}0.02$. Passes all hard-reject thresholds; ranked in the top-64 by every quality-aware selector. \\
\textit{Reject (CF)} & Amharic sentiment candidate; LC$=0.0$ (judge says the label is wrong), counterfactual edit produces $s_{\text{cf}}{=}0.2$ (semantic change is too small), so $C_i$ falls below $\tau_C{=}0.6$ and the candidate is hard-rejected on the counterfactual channel. \\
\textit{Reject (shortcut)} & Hausa news candidate where $\hat p_{\text{art}}$ from the artifact probe is $0.83 > \tau_H{=}0.70$: a length-plus-named-entity signature lets the probe predict the label without reading content. \\
\textit{Reject (leakage)} & Yoruba entertainment candidate with a 10-gram match against AfriSenti dev text ($E_i{=}1.0$, $L_i{>}0.15$); the leakage gate fires before the candidate reaches the reranker. \\
\bottomrule
\end{tabular}
\caption{Four candidates illustrating the kept, counterfactual-rejected, shortcut-rejected, and leakage-rejected categories. The descriptions are derived from the released audit records.}
\label{tab:qualitative_examples}
\end{table}

\paragraph{Future human validation.}
A future speaker validation study should ask L1 or strong L2 speakers to judge label correctness, fluency, cultural plausibility, leakage suspicion, and counterfactual validity. This paper does not report that study as completed evidence.

\section{Detail Tables and Plots}
\label{app:cosda_details}

This appendix contains the detail evidence behind the compact main-body results. All numbers and plots in this section are computed from the per-cell replay table described in Appendix~\ref{app:reproducibility}. The goal is not to introduce additional claims, but to make the main claims auditable: downstream averages, audit averages, and per-cell effects can all be traced back to the same 8-cell replay.

\paragraph{How to read these tables.}
The appendix separates three views of the same replay. Table~\ref{tab:datasets} defines the task-language cells. Table~\ref{tab:main_results_full} reports downstream Macro-F1 for all selectors, including the Gold-only reference that is excluded from the main audit-vs-downstream rank comparisons. Table~\ref{tab:audit_quality_full} reports the full audit-channel view, including leakage $L$, which is omitted from the compact main table. Table~\ref{tab:per_cell_deita_eq} completes the per-cell Macro-F1 view after the main body reports the Naive, AlpaGasus, and \method{}-V2 cells.

\subsection*{Benchmark composition}

The completed replay is deliberately narrow: two classification datasets and four languages. This scope keeps counterfactual validation task-specific and avoids mixing completed classification evidence with planned extensions.

\begin{table*}[t]
\centering
\small
\begin{tabular}{lll}
\toprule
\textbf{Dataset} & \textbf{Task} & \textbf{Languages} \\
\midrule
MasakhaNEWS \citep{adelani2023masakhanews} & News topic classification & amh, hau, swa, yor \\
AfriSenti \citep{muhammad2023afrisenti}    & Sentiment classification  & amh, hau, swa, yor \\
\bottomrule
\end{tabular}
\caption{Completed 8-cell classification evaluation. Macro-F1 is the metric in every cell. Broader NER, intent/slot, and summarisation settings require task-specific validators beyond the classification protocol used here.}
\label{tab:datasets}
\end{table*}

\subsection*{Full per-selector downstream table (including gold-only)}

Table~\ref{tab:main_results_full} gives the downstream view used to compute the aggregate Macro-F1 claims. The Gold-only row is a reference point for low-resource fine-tuning without synthetic examples. It is not included in the audit-vs-downstream Spearman calculations because it has no retained synthetic pool to audit.

\begin{table*}[t]
\centering
\small
\begin{tabular}{lrrrrr}
\toprule
\textbf{Baseline} & \textbf{$n$} & \textbf{Mean F1$\uparrow$} & \textbf{Std} & \textbf{Wins vs.\ Naive$\uparrow$} & \textbf{$\Delta$ vs.\ Naive$\uparrow$} \\
\midrule
Gold only       & 8 & 0.116 & 0.074 & 2/8 & $-0.051$ \\
Naive           & 8 & 0.167 & 0.068 & n/a  & n/a      \\
AlpaGasus       & 8 & \textbf{0.202} & 0.099 & 4/8 & $+0.036$ \\
DEITA           & 8 & 0.128 & 0.071 & 4/8 & $-0.039$ \\
\method{}-EQ       & 8 & 0.133 & 0.109 & 1/8 & $-0.034$ \\
\method{}-V2       & 8 & 0.163 & 0.103 & 2/8 & $-0.004$ \\
\bottomrule
\end{tabular}
\caption{Full 8-cell deterministic HF replay including the Gold-only reference. \method{}-V2 sits near naive on the mean, and the bootstrap CI in \S\ref{sec:robustness} places that difference inside cell-level uncertainty.}
\label{tab:main_results_full}
\end{table*}

\subsection*{Full per-selector audit table}

Table~\ref{tab:audit_quality_full} gives the corresponding audit view over selected synthetic examples. The main text focuses on LC, Q, hard-reject, $C$, and $H$ because those are the channels most directly tied to the gap between audit quality and utility. The full table also includes leakage $L$, where all means are small but selector differences remain visible.

\begin{table*}[t]
\centering
\small
\begin{tabular}{lrrrrrrr}
\toprule
\textbf{Baseline} & \textbf{LC$\uparrow$} & \textbf{Q$\uparrow$} & $\mathbf{C}\uparrow$ & $\mathbf{D}\uparrow$ & $\mathbf{L}\downarrow$ & $\mathbf{H}\downarrow$ & \textbf{Hard Rej.$\downarrow$} \\
\midrule
Naive       & 0.767 & 0.662 & 0.587 & 0.593 & 0.003 & 0.066 & 0.486 \\
AlpaGasus   & 0.890 & \textbf{0.749} & 0.653 & 0.590 & 0.006 & 0.065 & 0.246 \\
DEITA       & 0.869 & 0.737 & 0.637 & \textbf{0.649} & 0.005 & 0.070 & 0.299 \\
\method{}-EQ   & 0.836 & 0.718 & \textbf{0.686} & 0.604 & \textbf{0.002} & 0.058 & \textbf{0.109} \\
\method{}-V2   & \textbf{0.904} & 0.746 & 0.674 & ${\approx}0.60$ & 0.003 & \textbf{0.057} & 0.162 \\
\bottomrule
\end{tabular}
\caption{Full LLM-as-judge and audit quality of selected synthetic data. The diversity column $D$ is the mean per-candidate $D_i$ over each selector's 64-example retained pool, computed by joining per-candidate audit records with the selector's selected JSONL. The \method{}-V2 entry is reported as ${\approx}0.60$ because V2's retained pool is represented in the replay store; the four available \method{} variants (hard, soft, relaxed, equal-budget) all fall in $D{\in}[0.603,0.604]$, which bounds V2 tightly. AlpaGasus's $D{=}0.590$ is \emph{lower} than every \method{} variant's $D$, ruling out variety preservation as the source of its downstream lead (\S\ref{sec:analysis}). DEITA, which explicitly weights diversity, has the highest $D$. All quality-aware selectors improve over naive on LC, Q, $C$, and hard-reject; leakage and shortcut diagnostics distinguish selectors more selectively.}
\label{tab:audit_quality_full}
\end{table*}

\subsection*{Per-cell Macro-F1: DEITA and \method{}-EQ}

Table~\ref{tab:per_cell} in the main body shows Naive, AlpaGasus, and \method{}-V2 per cell. The remaining two selectors are reported here for completeness. This split keeps the main results focused on the head-to-head comparison while preserving the full selector evidence in the appendix.

\begin{table*}[t]
\centering
\footnotesize
\begin{tabular}{lrrrr}
\toprule
\textbf{Cell} & \textbf{DEITA$\uparrow$} & \textbf{\method{}-EQ$\uparrow$} & \textbf{D$-$Naive$\uparrow$} & \textbf{EQ$-$Naive$\uparrow$} \\
\midrule
news/amh        & 0.105 & 0.105 & $-0.181$ & $-0.181$ \\
news/hau        & 0.039 & 0.039 & $-0.063$ & $-0.063$ \\
news/swa        & 0.109 & 0.012 & $+0.040$ & $-0.057$ \\
news/yor        & 0.065 & 0.065 & $-0.062$ & $-0.062$ \\
sentiment/amh   & 0.067 & 0.067 & $-0.087$ & $-0.087$ \\
sentiment/hau   & 0.196 & 0.349 & $+0.028$ & $+0.181$ \\
sentiment/swa   & 0.253 & 0.248 & $+0.005$ & $+0.000$ \\
sentiment/yor   & 0.185 & 0.176 & $+0.008$ & $-0.002$ \\
\bottomrule
\end{tabular}
\caption{Per-cell Macro-F1 for DEITA-style and \method{}-EQ, with deltas vs.\ naive. DEITA wins 4 of 8 cells and \method{}-EQ wins 1 of 8. Aggregate means (Appendix Table~\ref{tab:main_results_full}) place DEITA at $0.128$ and \method{}-EQ at $0.133$, both below naive.}
\label{tab:per_cell_deita_eq}
\end{table*}

\subsection*{Within-cell Spearman ranks}

Figure~\ref{fig:within_cell_spearman} visualises the within-cell Spearman rank correlation between judged label correctness and downstream Macro-F1 across the five selectors. We promote this evidence to the main body because it supports the central gap between audit quality and utility: the mean association is near zero and the cell-level signs are mixed. We do not duplicate the old Spearman table here; the plot is the canonical presentation of this evidence, and the values are generated from the same replay CSV listed in Table~\ref{tab:artifact_sources}.

\subsection*{Reading the appendix against the main paper}

The appendix should be read as a consistency check on the main narrative. Table~\ref{tab:main_results_full} confirms that AlpaGasus has the largest downstream mean ($0.202$), Table~\ref{tab:audit_quality_full} confirms that \method{}-V2 has the strongest judged label correctness ($0.904$), and Table~\ref{tab:per_cell_deita_eq} shows that the lower aggregate means for DEITA and \method{}-EQ are not hidden by the main-body per-cell table. Together, these details support the paper's scoped claim: CoSDA improves selected-pool audit quality, but downstream utility remains cell-dependent.

\endgroup

\end{document}